\documentclass[conference]{IEEEtran}
\IEEEoverridecommandlockouts
\usepackage{cite}
\usepackage{amsmath,amssymb,amsfonts}
\usepackage{algorithmic}
\usepackage{graphicx}
\usepackage{textcomp}
\usepackage{xcolor}
\usepackage[disable]{todonotes}
\usepackage{balance}

\usepackage[utf8]{inputenc}
\usepackage[T1]{fontenc}
\usepackage{tabularx}
\usepackage{adjustbox}
\usepackage{array}
\usepackage{booktabs}

\usepackage{url}

\def\BibTeX{{\rm B\kern-.05em{\sc i\kern-.025em b}\kern-.08em
    T\kern-.1667em\lower.7ex\hbox{E}\kern-.125emX}}
\begin{document}

\title{Entity Extraction from High-Level Corruption Schemes via Large Language Models
\thanks{The research leading to these results has received funding from the European Union’s Internal Security Fund under grant agreement No 101103298 (KLEPTOTRACE). This publication reflects only the authors’ views. The European Commission is not responsible for any use that may be made of the information it contains. 
ChatGPT was utilized to generate sections of this work, including text, code, data, and citations.}
}

\author{\IEEEauthorblockN{Panagiotis Koletsis}
\IEEEauthorblockA{\textit{Dept. of Informatics and Telematics} \\
\textit{Harokopio University of Athens}\\
Athens, Greece \\
pkoletsis@hua.gr}
\and
\IEEEauthorblockN{Panagiotis-Konstantinos Gemos}
\IEEEauthorblockA{\textit{Dept. of Informatics and Telematics} \\
\textit{Harokopio University of Athens}\\
Athens, Greece \\
pgemos@hua.gr}
\and
\IEEEauthorblockN{Christos Chronis}
\IEEEauthorblockA{\textit{Dept. of Informatics and Telematics} \\
\textit{Harokopio University of Athens}\\
Athens, Greece \\
chronis@hua.gr}
\and
\IEEEauthorblockN{Iraklis Varlamis}
\IEEEauthorblockA{\textit{Dept. of Informatics and Telematics} \\
\textit{Harokopio University of Athens}\\
Athens, Greece \\
varlamis@hua.gr}
\and
\IEEEauthorblockN{Vasilis Efthymiou}
\IEEEauthorblockA{\textit{Dept. of Informatics and Telematics} \\
\textit{Harokopio University of Athens}\\
Athens, Greece \\
vefthym@hua.gr}
\and
\IEEEauthorblockN{Georgios Th. Papadopoulos}
\IEEEauthorblockA{\textit{Dept. of Informatics and Telematics} \\
\textit{Harokopio University of Athens}\\
Athens, Greece \\
g.th.papadopoulos@hua.gr}
}

\maketitle

\begin{abstract}
The rise of financial crime that has been observed in recent years has created an increasing concern around the topic and many people, organizations and governments are more and more frequently trying to combat it. Despite the increase of interest in this area, there is a lack of specialized datasets that can be used to train and evaluate works that try to tackle those problems. This article proposes a new micro-benchmark dataset for algorithms and models that identify individuals and organizations, and their multiple writings, in news articles, and presents an approach that assists in its creation. Experimental efforts are also reported, using this dataset, to identify individuals and organizations in financial-crime-related articles using various low-billion parameter Large Language Models (LLMs). For these experiments, standard metrics (Accuracy, Precision, Recall, F1 Score) are reported and various prompt variants comprising the best practices of prompt engineering are tested. In addition, to address the problem of ambiguous entity mentions, a simple, yet effective LLM-based disambiguation method is proposed, ensuring that the evaluation aligns with reality. Finally, the proposed approach is compared against a widely used state-of-the-art open-source baseline, showing the superiority of the proposed method.
\end{abstract}




\section{Introduction}\label{sec:intro}




Financial crimes have always been a concern, but the recent geopolitical instabilities have played a crucial role in a significant increase of illicit financial activities, including sanctions evasion, money laundering, and the use of shadowy financial networks to fund conflicts and circumvent international restrictions~\cite{de2024navigating}. Geopolitical tensions have created opportunities for criminals to exploit the chaos and for sanctioned individuals and entities to find new ways to access global markets~\cite{finelli2023countering}. In general, most research work in financial data is focused on areas related to the stock market, such as market analysis~\cite{ao2018sentiment} and stock price prediction~\cite{DBLP:conf/icacci/KuttichiraGMS17}. The primary goal of this paper is to extract information related to financial crime data, and contribute to the limited literature in the topic. The information extraction is conducted on news articles that mention individuals and organizations. This information can then be used by experts in the investigation of financial crime cases.

The latest advances in LLMs have sparked a new family of algorithms that rely on prompt engineering to perform natural-language processing (NLP) tasks, like named entity recognition (NER)~\cite{tsirmpas2024neural} or attribute value extraction~\cite{fang2024llm}. Therefore, a natural question that arises is how well would an LLM with appropriate prompting work in the tasks described above. Due to the lack of research in this specific field, there are no datasets specialized in financial crime that can be used to evaluate the performance of the LLM responses. In this case, the interest is specifically focused on a list of named entities, individuals (persons) or organizations, mentioned within some provided articles. To address this, an innovative pipeline for dataset generation has been created without compromising reliability. The resulting dataset contains ground truth for individual and organization identification in JSON format for each included article. In addition, a helper LLM is used to fix the format of malformed JSON responses. Lastly, the issue of ambiguous entity mentions, (i.e., references to the same element in different ways) is tackled by developing a new pipeline that uses an LLM to extract matching elements.

In summary, the contributions of this work are the following: 
\begin{itemize}
    \item A new Individual and Organization identification micro-benchmark dataset related to financial crimes (e.g., corruption, sanction evasion).
    \item A new LLM-based Individual and Organization identification method.
    \item A new LLM-based evaluation method for Individual and Organization identification.
\end{itemize}
The source code of this work is publicly available on GitHub: \url{https://github.com/panagiotis-koletsis/KleptotraceDataset}

\textbf{Overview.} The rest of the paper is organized as follows. Section~\ref{sec:related_work} overviews the related works and highlights the novel contributions of this paper. Section~\ref{sec:methodology}, provides the details of the developed methodology. Section~\ref{sec:experiments} presents the experimental evaluation of the proposed method. The paper is concluded in Section~\ref{sec:conclusion}.

\section{Related Work}\label{sec:related_work}

Research efforts on financial data mostly focus on stock market and other commodities price prediction and analysis. NLP tasks that aid these targets include sentiment analysis from news articles~\cite{mohan2019stock} and extracting information from complex company earning reports~\cite{glodd2023extraction}. There is, though, much work on financial data related to crime, with most publications focusing on fraud detection~\cite{hilal2022financial}. However, there is little research on the identification of parties related to financial crimes from a list of documents. Most papers performing NER for identification of entities of interest are not focusing on a particular form of crime. For example, in the work of Das et al.~\cite{das2020framework} data related to crime against women in India are used and in the work of Wu et al.~\cite{wu2020public} data related to public opinion monitoring are used.

In addition, very few works use LLMs to perform NER on crime-related data and more specifically on financial corruption data. One of the most notable ones is that of Hammami et al.~\cite{hammami2024fighting}. This work, however, only focuses on the identification and parsing of street addresses. The purpose is to locate the parties involved in financial transactions, since these are written in free text. Having identified the address of the parties can be later used for crime detection. The aforementioned work uses Transformers and LLMs for the identification process.

\textbf{Financial Market Tasks.} Most research regarding financial data is mainly focused in the area of financial markets, due to the wide range of interest that exists. This is true for both classical Machine Learning Methods and Deep Learning Methods~\cite{nosratabadi2020data}. Especially for LLMs the research in this area is currently very active. Starting with FinBERT~\cite{DBLP:conf/ijcai/0001HH0Z20}, a transformer model based on BERT, which primarily focuses on sentiment analysis of financial data, achieving state-of-the-art performance on related datasets, such as the Financial Phrase Bank in 2019~\cite{DBLP:journals/corr/abs-1908-10063,DBLP:journals/jasis/MaloSKWT14}. In 2023, Bloomberg launched its own LLM based on the BLOOM architecture, called BloombergGPT, an enterprise AI assistant for financial tasks such as investment decisions and risk management~\cite{DBLP:journals/corr/abs-2303-17564}. Shortly after, the need for an open-source version of BloombergGPT is met with the release of FinGPT, providing similar capabilities to the Bloomberg counterpart~\cite{DBLP:journals/corr/abs-2306-06031}.

\textbf{General Tasks in NER Datasets.}
NER has been a trending subject for the past few decades~\cite{roy2021recent}. Initially, statistical models provided acceptable results~\cite{naseer2021named}. More recently, a long short-term memory (LSTM)-based model called Flair embeddings~\cite{DBLP:conf/coling/AkbikBV18} achieved a 93.09 F1 score on the CoNLL-2003 Hugging Face dataset\footnote{\url{https://huggingface.co/datasets/eriktks/conll2003}}. With the introduction of the attention layer and the implementation of transformer models, results are further improved~\cite{vaswani2017attention}. A BERT-based model called LUKE achieved a 94.3 F1 score on the same dataset~\cite{DBLP:conf/emnlp/YamadaASTM20}. The state-of-the-art performance on this dataset is currently achieved not by a single model, but by a technique that uses reinforcement learning to search for the best embedding among multiple candidates provided by different models, named ACE~\cite{DBLP:conf/acl/WangJBWHHT20a}.
\section{Methodology}\label{sec:methodology}
In this section, a new micro-benchmark dataset generation method is proposed, related to financial corruption data. Additionally, the evaluation process seeks to identify entity mentions that are semantically (and not just lexicographically) equivalent. 

\begin{figure*}
    \centering
    \includegraphics[width=0.8\textwidth]{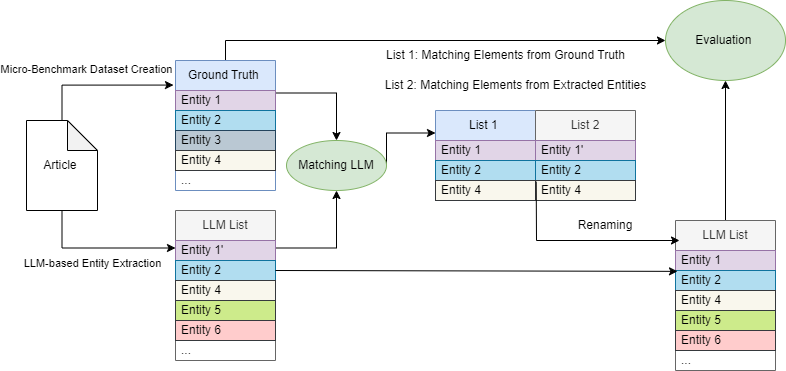}
    \caption{Overall architecture, including micro-benchmark generation and evaluation methodology.}
    \label{fig:full_pipeline}
\end{figure*}

The overall architecture is presented in (Figure~\ref{fig:full_pipeline}), while its components are presented in more detail in the following. 

\subsection{Micro-Benchmark Dataset Generation}\label{ssec:benchmark_generation}
\begin{figure}[h]
    \centering
    \includegraphics[width=0.5\textwidth]{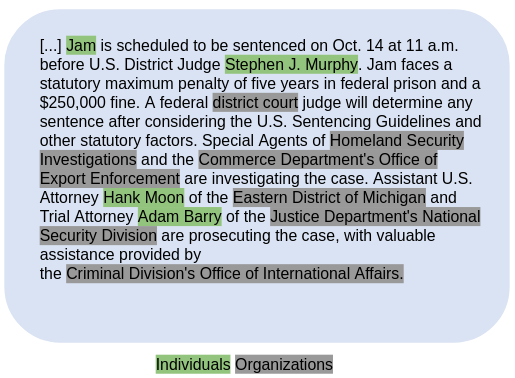}
    \caption{An example of entities detected in a snippet of a  given article.}
    \label{fig:Article_example}
\end{figure}

The benchmark generation methodology receives as input a set of articles and returns annotations, as lists for those articles, reflecting the named entities (individuals and organizations). In this use case, the benchmark generation pipeline takes as input a large dataset containing multiple articles from various cases, that is provided within the context of EU KLEPTOTRACE project~\cite{kleptotrace-project-2024}. This large unstructured dataset is comprised of more than 5000 multilingual articles from over 20 cases, collected from trusted sources, such as law enforcement press releases. An example is shown in (Figure~\ref{fig:Article_example}), highlighting the annotations (detected individuals and organizations) returned for a given article. For presentation purposes, only a snippet of this article is shown in the figure.

\subsubsection{Annotation Process} 
\begin{figure}
    \centering
    \includegraphics[width=0.8\columnwidth]{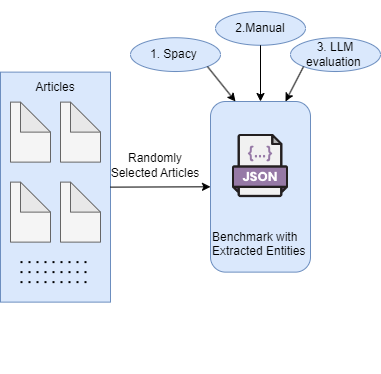}
    \vspace{-30pt}
    \caption{Micro-benchmark dataset creation.}
    \label{fig:Micro-Dataset}
\end{figure}

The annotation process involves the following steps, as shown in (Figure~\ref{fig:Micro-Dataset}):
\begin{itemize}
  \item spaCy annotation: The spaCy library is utilized to assist with manual annotation.
  \item Manual annotation: Inspect, correct and expand spaCy's annotations.
  \item Final verification by a LLM specifically ChatGPT 3.5. 
  \item To ensure accuracy in identifying the organizations, answers are validated through comprehensive web searches when needed.
\end{itemize}

The dataset is preprocessed as needed. For example, special characters unsuitable for JSON format are replaced with suitable alternatives, and escape characters (`\textbackslash n') are added where necessary.

\subsubsection{Data Statistics and Description}
The generated dataset contains 15 randomly selected articles from the provided set of articles, annotated in a JSON format. 

The articles vary in length from 170 to 1,467 words. Each article contains a different number of Individuals/Organizations to be identified, ranging from 1 to 12 for Individuals and 0 to 16 for Organizations. In total, there are 84 individuals and 128 organizations mentioned in the dataset.
 
Each article has different characteristics, allowing the proposed methodology to be tested under various circumstances.
The final form of the dataset consists of 441 sentences, 11,152 words, and 72,332 characters.

Based on these statistics, it can be concluded that the dataset comprises complex sentences pertaining to financial crimes, with an average sentence length of 25.3 words. In contrast, the CoNLL-2003 dataset consists of simpler, more general sentences with an average of 14.5 words. 
Lastly, one characteristic of the current methodology that is worth mentioning, is the exclusive use of general-purpose LLMs without any fine-tuning being applied to them.

In summary, the key differences of the generated dataset to well-known NER benchmark datasets (like CoNNL-2003) are the following:
\begin{itemize}
  \item \textbf{Specialized dataset} (Financial Corruption theme) in contrast to general.
  \item Dataset is focusing on \textbf{article per article} basis and not sentence per sentence. 
  \item More \textbf{complicated sentences} in contrast to simple ones.
  \item Usage of \textbf{untampered multi-purpose LLMs}, not fine-tuned on NER tasks.
\end{itemize}

The Dataset can be accessed at: 

\url{https://zenodo.org/records/14027005}

\subsection{Evaluation Methodology}\label{ssec:evaluation_method}
In order to evaluate a NER method on the generated benchmark data, a novel LLM-based evaluation methodology is proposed, considering not only lexicographical equivalence (e.g., exact matches), but also semantic similarity (e.g., synonyms, aliases). Even if the evaluation methodology is applied in the new benchmark dataset, it can also be applied to evaluated NER methods on other benchmark datasets. 

\begin{figure*}
    \centering
    \includegraphics[width=0.7\textwidth]{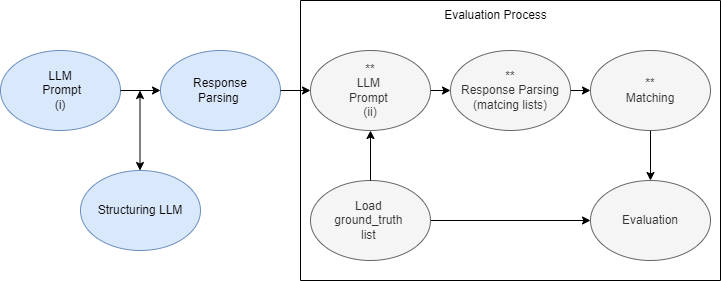}
    \caption{LLM-based extraction (left) and evaluation process (right). **Optional steps for organization identification.}
    \label{fig:methodology}
\end{figure*}

The LLM-based extraction process, as well as the evaluation process, are shown in Figure~\ref{fig:methodology} and described in detail, below. 

\textit{LLM Prompt (i)}: 
The base prompt development is achieved following a trial-and-error approach, ensuring that all necessary prompt engineering techniques are applied:
\begin{itemize}
  \item Specificity in Requirements: Clear specification of the requirements. 
  \item Response Format and Example: Detailed instructions on the desired response format are provided, along with an example to activate in-context learning.
\end{itemize}

\begin{figure}
    \centering
    \includegraphics[width=0.50\textwidth]{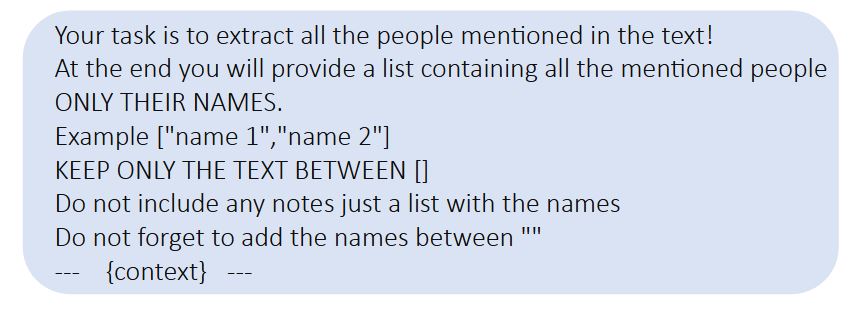}
    \caption{Base prompt for individual identification.}
    \label{fig:base_prompt_individual}
\end{figure}

\begin{figure}
    \centering
    \includegraphics[width=0.50\textwidth]{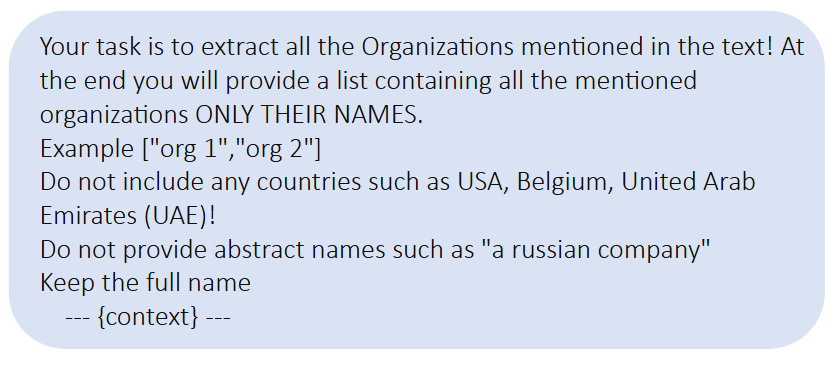}
    \caption{Base prompt for organization identification.}
    \label{fig:base_prompt_org}
\end{figure}

Two basic prompts are created, one for identifying individuals (Figure~\ref{fig:base_prompt_individual}) and one for identifying organizations (Figure~\ref{fig:base_prompt_org}). 

In the experiments section, additional best practices of prompt engineering are explored and their effectiveness in NER tasks analyzed.

\begin{figure}
    \centering
    \includegraphics[width=0.50\textwidth]{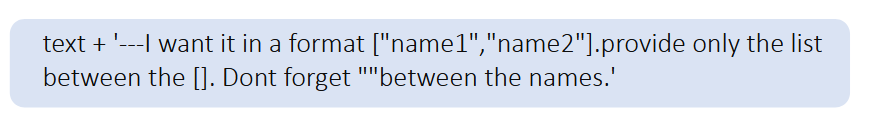}
    \caption{Prompt for structuring (Structuring LLM).}
    \label{fig:structuring}
\end{figure}

\textit{Structuring LLM}: Additionally, a helper prompt is developed to structure the response of the initial LLM (Figure~\ref{fig:structuring}). These prompts are located in their respective prompt templates.
Subsequently, each text (article) is iteratively passed through the prompt template to receive the LLM response. 

\textit{Response Parsing:} After obtaining the response, it is examined to ensure it adheres to the required patterns, thereby considering it valid. If the response is not considered valid, another layer of the LLM is activated. This layer utilizes the helper prompt, which ensures the correct structure of the response. 
Once a valid response is received, the useful part - specifically, the portion containing the list of all mentioned names (individuals or organizations) - is isolated, parsed and added to a list.

\begin{figure}
    \centering
    \includegraphics[width=0.50\textwidth]{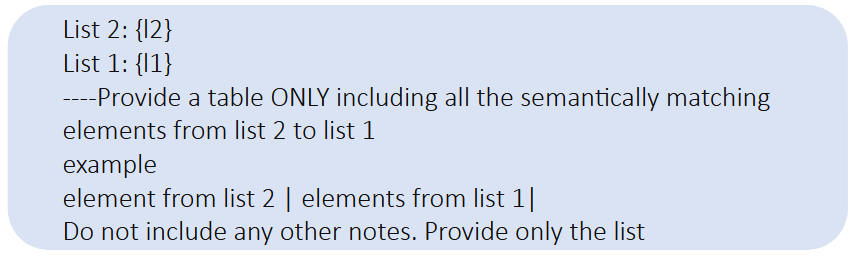}
    \caption{Prompt for matching elements (Matching LLM).}
    \label{fig:matchingPrompt}
\end{figure}

\textit{Matching LLM - LLM Prompt (ii)}: After obtaining the list of names, an optional step is available only for organization identification. This step addresses the inconsistency in the LLM's responses, which occurs due to the various ways the same organization can be referenced. For example, it is important to ensure that ``FBI'' will be considered the same as ``Federal Bureau of Investigations'', that ``General Office of International Affairs'' will be considered the same as ``Office of International Affairs'', etc.
To achieve this, the list created by the LLM and the list containing the Ground Truth are passed through a new prompt template, specifically designed to match the corresponding elements from the two lists (Figure~\ref{fig:matchingPrompt}). 

\begin{figure}
    \centering
    \includegraphics[width=0.50\textwidth]{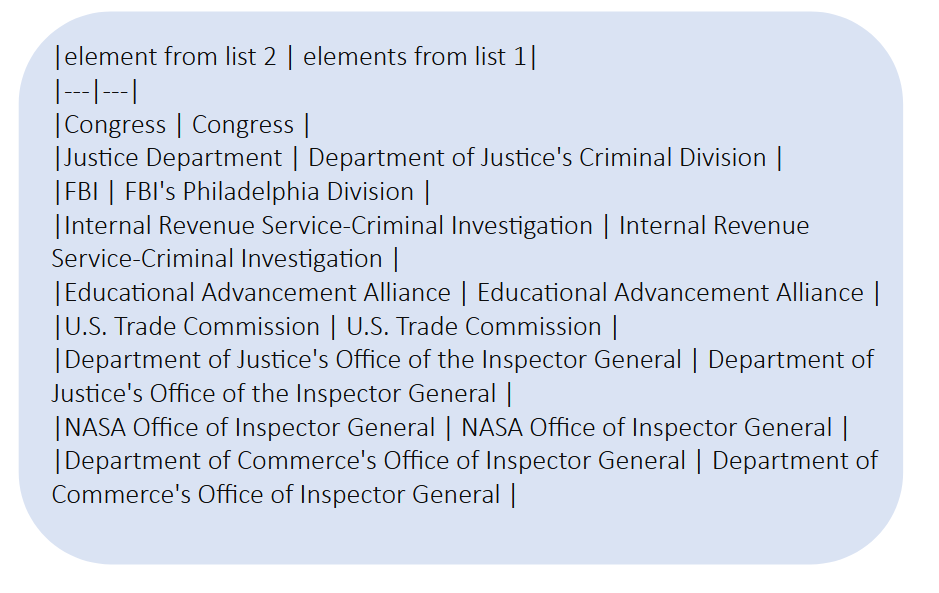}
    \caption{Example of matching LLM output.}
    \label{fig:matchingOutput}
\end{figure}

\textit{Matching LLM - Response Parsing \& Matching}: After processing, matching elements are isolated and parsed into two new lists: List 1 containing the elements from the Ground Truth annotations and List 2 containing the results from the LLM prompts. 
For example, the output of the matching prompt is in (Figure~\ref{fig:matchingOutput}). 
To summarize, there are four lists involved in this process (see Figure~\ref{fig:full_pipeline}):
\begin{itemize}
  \item Ground Truth: A list containing the results generated via the Micro-Benchmark Dataset Generation method (Section~\ref{ssec:benchmark_generation}).
  \item LLM List: A list containing the entities extracted from the LLM prompts (Figures~\ref{fig:base_prompt_individual} and \ref{fig:base_prompt_org}).
  \item List 1: A list containing the elements from the Ground Truth list that match an element from the LLM List, maintaining positional correspondences.
  \item List 2: A list containing the elements from the LLM List that match an element from the Ground Truth list, again maintaining positional correspondences.
\end{itemize}

Next, the matching elements from  List 1 and List 2 are renamed in LLM list, if needed, to use the same names as those used in List 1. (see Renaming in Figure~\ref{fig:full_pipeline}).
Finally, the updated LLM List is taken, which contains the initial elements from LLM List, with the common elements renamed as they appear in the Ground Truth list. 

\textit{Evaluation metrics:}
The similarity between the updated LLM List and the Ground Truth list is then evaluated, using standard metrics, such as accuracy, precision, recall, and F1-score. The renaming process described above ensures that the matching elements are considered identical (i.e., True Positives), despite different naming conventions/variations that may have been used between the evaluated NER method and the Ground Truth.

An overall diagram of the process is depicted in (Figure~\ref{fig:full_pipeline}), where it is shown that the Ground Truth contains Entity 1 while the LLM refers to the same entity as Entity 1'. The matching process aligns these two elements (shown in light purple), and in the final list, Entity 1 is being replaced and referenced the same way as in the Ground Truth list, thereby making it an exact match.

\section{Experiments}\label{sec:experiments}
In this section, the experimental setup is first described, followed by an analysis of the results.

\subsection{Experimental Setting}

\textbf{Dataset:}
The dataset used is the one referenced in section [III.A], which contains articles related to financial crimes and all the mentioned individuals and organizations.

\textbf{Hardware-Software:}
All the experiments were performed on a single GPU (RTX 3080 Ti with 12GB VRAM) and an i7-13700K processor with 64GB of RAM. The prompt development process and initial testing were conducted using OpenWebUI, an open-source, easy-to-use user interface for the Ollama models. The UI was running on Docker and was utilized to expedite the early stages of the process.

\textbf{Models:}
The focus was primarily on experimenting with the latest low-billion-parameter models from the Ollama library\footnote{\url{https://ollama.com/library}}, specifically those under 10 billion parameters (i.e., Gemma2:9b, Llama3:8b, Gemma:7b, Qwen2:7B, Mistral, Llama2:7b, Mixtral:8x7b, Zephyr). However, some medium-parameter models ranging from 14 billion to 27 billion parameters (Phi3:medium, Gemma2:27b) were also evaluated.
It is important to mention that Mistral, LLaMA2:7B, Mixtral:8x7B, Zephyr, Phi3:medium, Qwen2:7B, and Gemma2:27B were excluded from further testing due to their unsatisfactory preliminary results.

\textbf{Evaluation measures:}
In the evaluation, standard NER evaluation (as a classification task) metrics were utilized, including 
Precision, Recall, F1 and Accuracy.
Iteration time represents the time in seconds required for a full iteration of data, meaning all 15 articles are parsed through the architecture one time.
All the roundings for these numbers were done to the third decimal place. Iteration time was also tracked and rounded to the first decimal place. Lastly, the percentages mentioned were calculated using the following formula: 
\[
\text{percentage} = \left( \frac{(\text{final} - \text{initial})}{\text{initial}} \right )\times 100
\]

\begin{figure}
    \centering
    \includegraphics[width=0.50\textwidth]{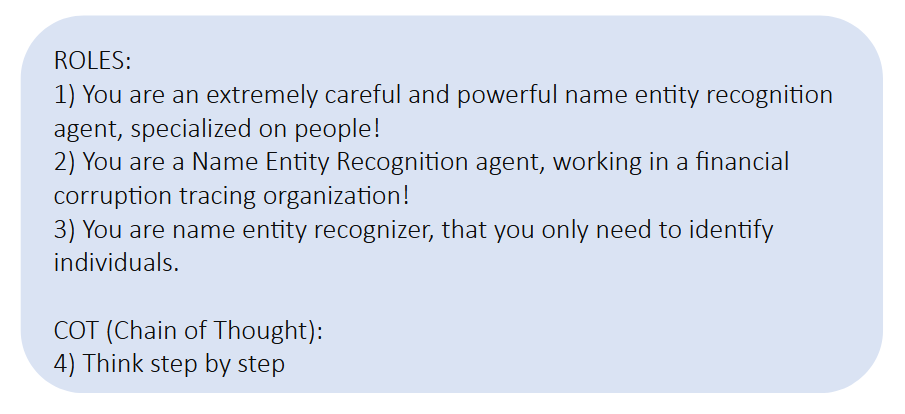}
    \caption{Prompt engineering on individual identification.}
    \label{fig:PromptEngInv}
\end{figure}

\begin{figure}
    \centering
    \includegraphics[width=0.50\textwidth]{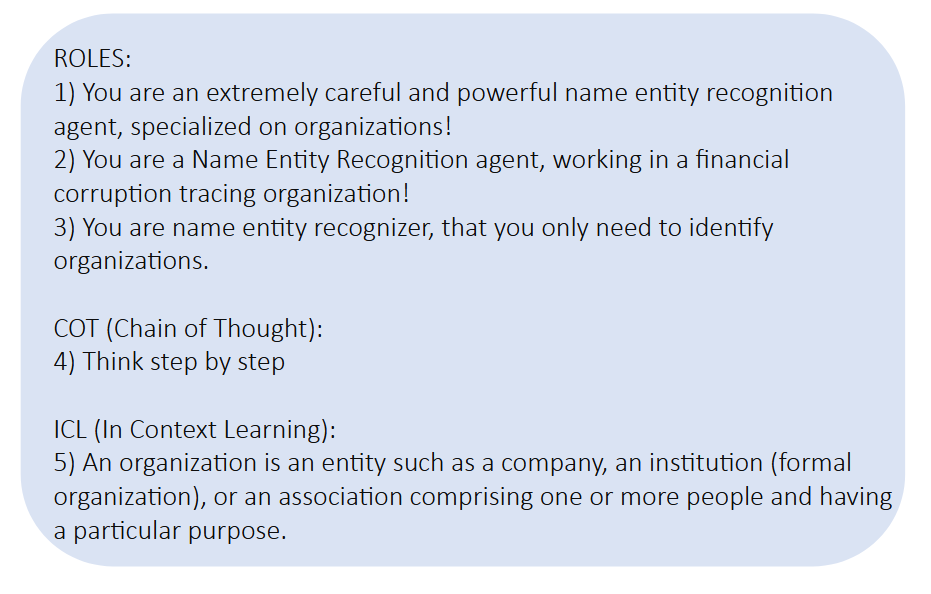}
    \caption{Prompt engineering on organization identification.}
    \label{fig:PrompEngOrg}
\end{figure}

\begin{figure}
    \centering
    \includegraphics[width=0.50\textwidth]{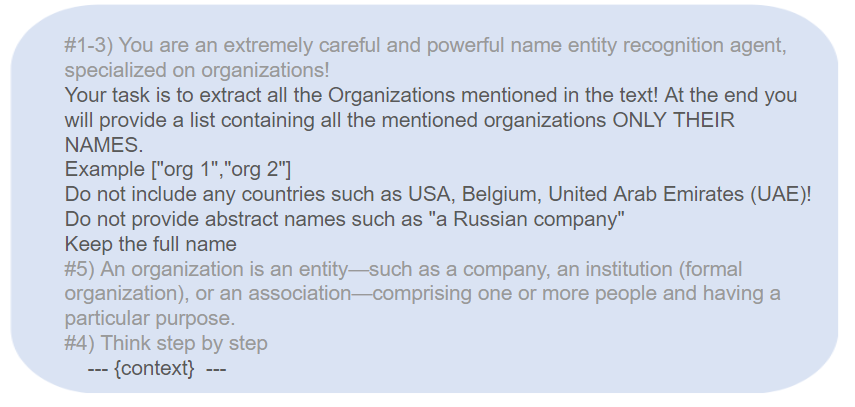}
    \caption{Example of full prompt.}
    \label{fig:full_prompt}
\end{figure}

\textbf{Baseline methods / Ablation study:}
After reviewing the base prompt for the task, the goal was to observe the differences with some revised prompts. In these revised prompts (refer to Figures~\ref{fig:PromptEngInv} and~\ref{fig:PrompEngOrg}), various best practices of prompt engineering were evaluated~\cite{DBLP:journals/corr/abs-2310-14735}. These sentences were added to the prompt to evaluate if there was any improvement in the evaluation score. Each modification was incorporated separately, and some combinations were also tested to assess for further improvements. 
These techniques include:
\begin{itemize}
  \item Roles: Asking the model to assume a specific role, helps it to adopt to a particular perspective. Three different roles were tested for each (Figures~\ref{fig:PromptEngInv} and \ref{fig:PrompEngOrg}).
  \item Chain of Thought: By specifically instructing the model to think step-by-step, the  aim was to improve its reasoning capabilities (Figure~\ref{fig:full_prompt}).
  \item Context: Providing additional context activates the model's ability to learn new information, a technique also known as In-Context Learning (ICL) (Figure~\ref{fig:full_prompt}).
\end{itemize}
Some additional best practices for prompt engineering were incorporated into the base prompt, including formatting the response, providing examples (also activates ICL)~\cite{DBLP:journals/corr/abs-2301-00234}, and being specific. Finally, the results are compared to a well-established baseline NER method, spaCy\footnote{\url{https://spacy.io/}}.

\subsection{Experimental Results}

In this section, we report the evaluation results on the NER task using the LLM prompts described in Section~\ref{ssec:evaluation_method}, on the new benchmark dataset described in Section~\ref{ssec:benchmark_generation}. 
More specifically, in Tables~\ref{tab:experimentInd} and~\ref{tab:experimentsOrg}, we present the results of Individual identification and Organization identification, respectively, with the "Prompt Addition" column showing the additions made to the base prompt (shown in Figures~\ref{fig:PromptEngInv} and ~\ref{fig:PrompEngOrg}). These tables also present a one-to-one comparison with the spaCy framework.  
Finally in Table~\ref{tab:individual_structuring_errors} we are discussing the results of the Structuring LLM component.

\begin{table*}[ht]
    \centering
    \caption{Individual identification results (Figure~\ref{fig:PromptEngInv}).}
    \renewcommand{\arraystretch}{1.3}  
    \begin{tabular}{@{}lcccccc@{}}  
        \toprule
        \textbf{Base model} & \textbf{Accuracy} & \textbf{Precision} & \textbf{Recall} & \textbf{F1 Score} & \textbf{Execution Time (Sec)} & \textbf{Prompt Additions} \\
        \midrule
        qwen2:7b & 0.523 & 0.862 & 0.570 & 0.686 & 10.1 & - \\
        llama3:8b & 0.882 & 0.963 & 0.913 & 0.937 & 12.3 & - \\
        llama3:8b & 0.861 & 0.958 & 0.895 & 0.925 & 12.5 & 1) \\
        llama3:8b & 0.865 & 0.965 & 0.893 & 0.927 & 12.3 & 2) \\
        llama3:8b & 0.820 & 0.956 & 0.852 & 0.901 & 11.6 & 3) \\
        llama3:8b & 0.850 & 0.957 & 0.883 & 0.918 & 12.0 & 4) \\
        gemma:7b & 0.859 & 0.918 & 0.930 & 0.924 & 67.6 & - \\
        gemma:7b & 0.818 & 0.908 & 0.891 & 0.899 & 66.6 & 1) \\
        gemma:7b & 0.865 & 0.931 & 0.923 & 0.927 & 67.0 & 2) \\
        gemma:7b & 0.836 & 0.933 & 0.889 & 0.910 & 65.0 & 3) \\
        gemma:7b & 0.854 & 0.910 & 0.932 & 0.921 & 67.6 & 4) \\
        gemma2:9b & 0.935 & 0.982 & 0.951 & 0.966 & 14.0 & - \\
        gemma2:9b & 0.858 & 0.927 & 0.920 & 0.923 & 14.4 & 1) \\
        gemma2:9b & 0.942 & 0.959 & \textbf{0.981} & 0.970 & 14.4 & 2) \\
        gemma2:9b & 0.942 & 0.978 & 0.963 & 0.970 & 14.2 & 3) \\
        \textbf{gemma2:9b} & \textbf{0.958} & \textbf{0.986} & 0.971 & \textbf{0.978} & 14.2 & \textbf{4)} \\
        gemma2:9b & 0.937 & 0.954 & \textbf{0.981} & 0.967 & 14.5 & 2),4) \\
        gemma2:9b & 0.867 & 0.942 & 0.917 & 0.929 & 14.2 & 1),4) \\
        spaCy & 0.447 & 0.486 & 0.845 & 0.617 & \textbf{0.9}  & - \\
        \bottomrule
    \end{tabular}
    \label{tab:experimentInd}
\end{table*}

\begin{table*}[ht]
    \centering
    \caption{Organization identification results (Figure~\ref{fig:PrompEngOrg}).}
    \renewcommand{\arraystretch}{1.3}  
    \begin{tabular}{@{}lccccccc@{}}  
        \toprule
        \textbf{Base model} & \textbf{Accuracy} & \textbf{Precision} & \textbf{Recall} & \textbf{F1 Score} & \textbf{Execution Time (Sec)} & \textbf{Prompt Additions} & \textbf{Matching} \\
        \midrule
        qwen2:7b & 0.342 & 0.764 & 0.382 & 0.508 & 74.9 & - & yes \\
        llama3:8b & 0.514 & 0.851 & 0.565 & 0.678 & 83.5 & - & yes \\
        gemma:7b & 0.523 & 0.767 & 0.622 & 0.687 & 114.5 & - & yes\\
        gemma2:9b & 0.640 & \textbf{0.961} & 0.657 & 0.780 & \textbf{36.2} & - & yes\\
        gemma2:9b & 0.663 & 0.941 & 0.692 & 0.797 & 37.6 & 1) & yes \\
        gemma2:9b & 0.636 & 0.914 & 0.676 & 0.777 & 37.2 & 2) & yes\\
        gemma2:9b & 0.612 & 0.913 & 0.650 & 0.759 & 36.5 & 3) & yes\\
        gemma2:9b & 0.683 & 0.949 & 0.709 & 0.811 & 48.8 & 4) & yes\\
        gemma2:9b & 0.653 & 0.946 & 0.679 & 0.790 & 37.9 & 5) & yes\\
        \textbf{gemma2:9b} & \textbf{0.699} & 0.936 & \textbf{0.734} & \textbf{0.823} & 59.5 & \textbf{1),4),5)} & yes\\
        spaCy & 0.117 & 0.176 & 0.258 & 0.209 & \textbf{0.9} &  - & no\\
        spaCy & 0.297 & 0.393 & 0.549 & 0.458 & 27.7 &  - & yes \\
        \bottomrule
    \end{tabular}
    \label{tab:experimentsOrg}
\end{table*}

\begin{table}[ht]
    \centering
    \caption{Individual identification for Structuring LLM.}
    \renewcommand{\arraystretch}{1.3}  
    \begin{tabular}{@{}lccc@{}}  
        \toprule
        \textbf{Model} & \textbf{Json errors} & \textbf{Iterations} & \textbf{Failure percent} \\
        \midrule
        \textbf{qwen2:7b} & \textbf{0} & \textbf{525} & \textbf{0\%} \\
        gemma2:9b & 4 & 525 & 0.76\% \\
        \bottomrule
    \end{tabular}
    \label{tab:individual_structuring_errors}
\end{table}


\textbf{Individual Identification:}
For this experiment, the focus was on LLaMA3:8B, Gemma:7B, and Gemma2:9b. It is important to note that these experiments started before the release of Gemma2:9B. Including Gemma2:9B was particularly gratifying, as the improvements are noticeable (Table~\ref{tab:experimentInd}).
\begin{itemize}
    \item llama3:8b
    \begin{itemize}
        \item It was unexpected that prompt engineering would fail to improve any of the tracked metrics. Despite having, one of the shortest iteration times, Gemma2:9B is considered a much better option.
    \end{itemize}
    \item gemma:7b
    \begin{itemize}
        \item The empirical observation that this model struggles with the structure of its responses was confirmed, as the structuring LLM was activated almost every time, thereby increasing the iteration time. Although slight improvements were achieved through prompt engineering (prompt additions) in some cases, these improvements were not consistent across all experiments. Nevertheless, the effects of prompt engineering were observed, achieving slight improvements for each tracked metric in various iterations.
    \end{itemize}
    \item gemma2:9b
    \begin{itemize}
        \item A slight performance improvement was achieved with prompt engineering. The improvement was slightly likely due to the straightforward nature of the task.
        \item A precision of 0.986 was achieved using the Chain of Thought technique (Prompt Addition: 4). During the same iteration, the highest accuracy and F1 score 0.978 was also achieved, with an increase of 1.42\% in iteration time.
        \item The combination of prompt additions that initially improved the F1 score did not lead to further improvement, highlighting the need for caution when modifying prompts.
    \end{itemize}
\end{itemize}

\textbf{Organization Identification:}
After running different variations of the base prompt using the suggested methodology for organization identification with Gemma2:9B, several key observations were made from Table~\ref{tab:experimentsOrg}:
\begin{itemize}
  \item Using a role that suits the model can boost the F1 score (Prompt Additions: 1-3), otherwise using a role could even decrease the F1 score.
  \item Providing context, even though the definition of "Organization" is well-known (Prompt Addition:5) and the model understands most of the information defining an organization, can also enhance the F1 score.
  \item The most significant improvement in performance was achieved through the Chain of Thought (Prompt Addition: 4) technique, which increased the F1 score by 3.97\%. However, this came with a performance cost of 34,80\%, as each iteration increased from 36.2 to 48.8 seconds.
  \item The best precision was achieved with the zero-shot (base prompt) approach, but it also resulted in one of the worst recalls. Additionally, it had the lowest iteration time among the LLM approaches.
  \item The combination of all prompt variations that boosted the F1 score yielded the highest F1 score while simultaneously having the worst performance in terms of time. This combination increased the F1 score by 5.51\% but also increased the iteration time by 64.36\%. Additionally, this combination achieved the best recall, improving it by 11.72\% and the best accuracy.
  \item It is clear that even slight improvements to the prompt come with a significant increase in iteration cost.
  \item 
  For the other two models, the LLM responsible for matching the elements from the two lists was Gemma2:9B, as the other models struggled with this task.
  \item 
  The other two models were tested only with the base prompt since their initial results were not competitive compared to the Gemma2:9B model. Significantly increased iteration times were noticed for these models, as Gemma2:9B's superior structuring capabilities contributed to its better performance. Consequently, the LLM responsible for structuring the first list gets activated more frequently, further increasing the iteration time.
\end{itemize}

\textbf{SpaCy comparison:}
The proposed methodology was compared with spaCy (Tables~\ref{tab:experimentInd} and~\ref{tab:experimentsOrg}), as it was employed directly in the dataset creation process. Spacy is a framework commonly used in industrial NLP processes. The \texttt{en\_core\_web\_sm} model from spaCy was specifically used and tested on the developed dataset to compare results. First, the spaCy framework with the raw model outputs was evaluated, and as expected, the results from the new proposed method were superior. Organization identification was also tested with the spaCy framework, including the matching component, and found that the suggested approach still outperformed spaCy. One notable advantage of using spaCy is its rapid output. In less than one second, spaCy was able to process all 15 of articles.

\textbf{Structuring results:}
To evaluate the Structuring LLM, the entire individual recognition experiment was ran 5 times, counting JSON errors and failure percetange, as shown in Table~\ref{tab:individual_structuring_errors}. This equates to 5 x 7 x 15 = 525 iterations, because each experiment involved 7 iterations of parsing each article, and there were 15 articles in total. Gemma2:9b had 4 JSON errors, indicating that 4 out of the 525 responses were not structured properly. However, in the conducted experiments, Qwen2:7B achieved perfect structuring with 0 JSON errors.

\textbf{Discussion}: 
Overall, one interesting finding was that both Gemma2:27B and Phi3 (which has 14 billion parameters), despite being recent and powerfull models in other NLP tasks, seemed to struggle with these tasks, as much as other, older models. 
Another observation was that, despite the initial unsatisfactory results of Qwen2:7B, zero JSON errors were encountered consistently with this model. This indicates that Qwen2:7B understood the structure of the format better than other models, such as LLaMA3:8B and Gemma:7B, which struggled significantly more. Therefore, Qwen2:7B  was used as the  Structuring LLM. 
However, Gemma2:9B outperformed other models (LLaMA3:8B, Gemma:7B) in structuring,which was empirically proven. Nevertheless, it did not match the performance of Qwen2:7B, which excelled at structuring its responses, as demonstrated experimentally (Table~\ref{tab:individual_structuring_errors}).

\textbf{Limitations}: Most of the limitations identified are resulting from the utilization of LLMs. 
\begin{itemize}
  \item Relatively slow and GPU intensive process.
  \item Suffering from hallucinations, which can affect the results and their trustworthiness.
  \item The results are non-deterministic, as the output of LLMs may vary across different iterations.
  \item Depending on third parties libraries for the LLM utilization, such as Ollama and Langchain. 
\end{itemize}

\section{Conclusion and Future Work}\label{sec:conclusion}

This work primarily focuses on identifying individuals and organizations in financial crime-related articles, as analyzing these cases is a highly relevant and timely topic. To achieve this, low-billion-parameter LLMs were used, and the proposed methodology was applied for dataset creation, structuring, and matching ambiguous entity mentions. The experimental results indicate the superiority of qwen2:7b on structuring its responses, therefore making it the structuring LLM. Later noticing that the best results were achieved by the Gemma2:9b model, these outcomes were further improved using prompt engineering best practices. Specifically, three key strategies were focused on: defining roles, incorporating  Chain of Thought, and adding extra context to the prompt. In this case, the Chain of Thought technique had the best impact on the result. Lastly, a direct comparison with SpaCy was conducted to prove the dominance of the suggested method.

One unexpected result that arose during empirical evaluation of various LLMs for integration with this methodology, was that medium billion-parameter models (14B - 27B) achieve worse results compared to low billion-parameter models (less than 10B). This observation requires further investigation to be generalized. However, based on the conducted experiments thus far, this trend was consistently noticed.

Finally, another interesting observation is that low-billion-parameter models, such as LLaMA3:8B and Mistral, generally did not improve scores on easy tasks such as individual identification and organization identification. This phenomenon also warrants further investigation.

\balance
\bibliographystyle{IEEEtran}
\bibliography{refs}

\begin{thebibliography}{10}
\providecommand{\url}[1]{#1}
\csname url@samestyle\endcsname
\providecommand{\newblock}{\relax}
\providecommand{\bibinfo}[2]{#2}
\providecommand{\BIBentrySTDinterwordspacing}{\spaceskip=0pt\relax}
\providecommand{\BIBentryALTinterwordstretchfactor}{4}
\providecommand{\BIBentryALTinterwordspacing}{\spaceskip=\fontdimen2\font plus
\BIBentryALTinterwordstretchfactor\fontdimen3\font minus \fontdimen4\font\relax}
\providecommand{\BIBforeignlanguage}[2]{{%
\expandafter\ifx\csname l@#1\endcsname\relax
\typeout{** WARNING: IEEEtran.bst: No hyphenation pattern has been}%
\typeout{** loaded for the language `#1'. Using the pattern for}%
\typeout{** the default language instead.}%
\else
\language=\csname l@#1\endcsname
\fi
#2}}
\providecommand{\BIBdecl}{\relax}
\BIBdecl

\bibitem{de2024navigating}
R.~De~Sybel, ``Navigating the storm: The intersection of geopolitical and financial crime risks,'' \emph{Journal of Risk Management in Financial Institutions}, vol.~17, no.~3, pp. 294--302, 2024.

\bibitem{finelli2023countering}
F.~Finelli, ``Countering circumvention of restrictive measures: The eu response,'' \emph{Common Market Law Review}, vol.~60, no.~3, 2023.

\bibitem{ao2018sentiment}
S.~Ao, ``Sentiment analysis based on financial tweets and market information,'' in \emph{ICALIP}.\hskip 1em plus 0.5em minus 0.4em\relax IEEE, 2018, pp. 321--326.

\bibitem{DBLP:conf/icacci/KuttichiraGMS17}
D.~P. Kuttichira, E.~A. Gopalakrishnan, V.~K. Menon, and K.~P. Soman, ``Stock price prediction using dynamic mode decomposition,'' in \emph{2017 International Conference on Advances in Computing, Communications and Informatics, {ICACCI} 2017, Udupi (Near Mangalore), India, September 13-16, 2017}, 2017, pp. 55--60.

\bibitem{tsirmpas2024neural}
D.~Tsirmpas, I.~Gkionis, G.~T. Papadopoulos, and I.~Mademlis, ``Neural natural language processing for long texts: A survey on classification and summarization,'' \emph{Engineering Applications of Artificial Intelligence}, vol. 133, p. 108231, 2024.

\bibitem{fang2024llm}
C.~Fang, X.~Li, Z.~Fan, J.~Xu, K.~Nag, E.~Korpeoglu, S.~Kumar, and K.~Achan, ``Llm-ensemble: Optimal large language model ensemble method for e-commerce product attribute value extraction,'' in \emph{Proceedings of the 47th International ACM SIGIR Conference on Research and Development in Information Retrieval}, 2024, pp. 2910--2914.

\bibitem{mohan2019stock}
S.~Mohan, S.~Mullapudi, S.~Sammeta, P.~Vijayvergia, and D.~C. Anastasiu, ``Stock price prediction using news sentiment analysis,'' in \emph{2019 IEEE fifth international conference on big data computing service and applications (BigDataService)}.\hskip 1em plus 0.5em minus 0.4em\relax IEEE, 2019, pp. 205--208.

\bibitem{glodd2023extraction}
A.~Glodd and D.~Hristova, ``Extraction of forward-looking financial information for stock price prediction from annual reports using nlp techniques,'' 2023.

\bibitem{hilal2022financial}
W.~Hilal, S.~A. Gadsden, and J.~Yawney, ``Financial fraud: a review of anomaly detection techniques and recent advances,'' \emph{Expert systems With applications}, vol. 193, p. 116429, 2022.

\bibitem{das2020framework}
P.~Das, A.~K. Das, J.~Nayak, and D.~Pelusi, ``A framework for crime data analysis using relationship among named entities,'' \emph{Neural Computing and Applications}, vol.~32, no.~12, pp. 7671--7689, 2020.

\bibitem{wu2020public}
W.~Wu, K.-P. Chow, Y.~Mai, and J.~Zhang, ``Public opinion monitoring for proactive crime detection using named entity recognition,'' in \emph{Advances in Digital Forensics XVI: 16th IFIP WG 11.9 International Conference, New Delhi, India, January 6--8, 2020, Revised Selected Papers 16}.\hskip 1em plus 0.5em minus 0.4em\relax Springer, 2020, pp. 203--214.

\bibitem{hammami2024fighting}
H.~Hammami, L.~Baligand, and B.~Petrovski, ``Fighting crime with transformers: Empirical analysis of address parsing methods in payment data,'' \emph{arXiv preprint arXiv:2404.05632}, 2024.

\bibitem{nosratabadi2020data}
S.~Nosratabadi, A.~Mosavi, P.~Duan, P.~Ghamisi, F.~Filip, S.~S. Band, U.~Reuter, J.~Gama, and A.~H. Gandomi, ``Data science in economics: comprehensive review of advanced machine learning and deep learning methods,'' \emph{Mathematics}, vol.~8, no.~10, p. 1799, 2020.

\bibitem{DBLP:conf/ijcai/0001HH0Z20}
Z.~Liu, D.~Huang, K.~Huang, Z.~Li, and J.~Zhao, ``Finbert: {A} pre-trained financial language representation model for financial text mining,'' in \emph{{IJCAI}}, 2020, pp. 4513--4519.

\bibitem{DBLP:journals/corr/abs-1908-10063}
D.~Araci, ``Finbert: Financial sentiment analysis with pre-trained language models,'' \emph{CoRR}, vol. abs/1908.10063, 2019.

\bibitem{DBLP:journals/jasis/MaloSKWT14}
P.~Malo, A.~Sinha, P.~J. Korhonen, J.~Wallenius, and P.~Takala, ``Good debt or bad debt: Detecting semantic orientations in economic texts,'' \emph{J. Assoc. Inf. Sci. Technol.}, vol.~65, no.~4, pp. 782--796, 2014.

\bibitem{DBLP:journals/corr/abs-2303-17564}
S.~Wu, O.~Irsoy, S.~Lu, V.~Dabravolski, M.~Dredze, S.~Gehrmann, P.~Kambadur, D.~S. Rosenberg, and G.~Mann, ``Bloomberggpt: {A} large language model for finance,'' \emph{CoRR}, vol. abs/2303.17564, 2023.

\bibitem{DBLP:journals/corr/abs-2306-06031}
H.~Yang, X.~Liu, and C.~D. Wang, ``Fingpt: Open-source financial large language models,'' \emph{CoRR}, vol. abs/2306.06031, 2023.

\bibitem{roy2021recent}
A.~Roy, ``Recent trends in named entity recognition (ner),'' \emph{arXiv preprint arXiv:2101.11420}, 2021.

\bibitem{naseer2021named}
S.~Naseer, M.~M. Ghafoor, S.~bin Khalid~Alvi, A.~Kiran, S.~U. Rahmand, G.~Murtazae, and G.~Murtaza, ``Named entity recognition (ner) in nlp techniques, tools accuracy and performance.'' \emph{Pakistan Journal of Multidisciplinary Research}, vol.~2, no.~2, pp. 293--308, 2021.

\bibitem{DBLP:conf/coling/AkbikBV18}
A.~Akbik, D.~Blythe, and R.~Vollgraf, ``Contextual string embeddings for sequence labeling,'' in \emph{{COLING}}, 2018, pp. 1638--1649.

\bibitem{vaswani2017attention}
A.~Vaswani, ``Attention is all you need,'' \emph{Advances in Neural Information Processing Systems}, 2017.

\bibitem{DBLP:conf/emnlp/YamadaASTM20}
I.~Yamada, A.~Asai, H.~Shindo, H.~Takeda, and Y.~Matsumoto, ``{LUKE:} deep contextualized entity representations with entity-aware self-attention,'' in \emph{{EMNLP}}, 2020, pp. 6442--6454.

\bibitem{DBLP:conf/acl/WangJBWHHT20a}
X.~Wang, Y.~Jiang, N.~Bach, T.~Wang, Z.~Huang, F.~Huang, and K.~Tu, ``Automated concatenation of embeddings for structured prediction,'' in \emph{{ACL/IJCNLP}}, 2021, pp. 2643--2660.

\bibitem{kleptotrace-project-2024}
\BIBentryALTinterwordspacing
K.~Project, ``{KLEPTOTRACE project - Co-funded by the European Commission},'' 7 2024. [Online]. Available: \url{https://transcrime.it/kleptotrace/}
\BIBentrySTDinterwordspacing

\bibitem{DBLP:journals/corr/abs-2310-14735}
B.~Chen, Z.~Zhang, N.~Langren{\'{e}}, and S.~Zhu, ``Unleashing the potential of prompt engineering in large language models: a comprehensive review,'' \emph{CoRR}, vol. abs/2310.14735, 2023.

\bibitem{DBLP:journals/corr/abs-2301-00234}
Q.~Dong, L.~Li, D.~Dai, C.~Zheng, Z.~Wu, B.~Chang, X.~Sun, J.~Xu, L.~Li, and Z.~Sui, ``A survey for in-context learning,'' \emph{CoRR}, vol. abs/2301.00234, 2023.

\end{thebibliography}

\end{document}